\documentclass[fleqn,10pt]{wlscirep}
\usepackage[utf8]{inputenc}
\usepackage[T1]{fontenc}
\title{On the dynamic evolution of CLIP texture–shape bias and its relationship to human alignment and model robustness}

\author[1,*,+]{Pablo Hernández-Cámara}
\author[2,+]{Jose Manuel Jaén-Lorites}
\author[3,4]{Alexandra Gómez-Villa}
\author[1]{Jorge Vila-Tomás}
\author[1]{Valero Laparra}
\author[1]{Jesús Malo}
\affil[1]{Image Processing Lab, Universitat de Valencia, Spain}
\affil[2]{Centro de Biomateriales e Ingenieria Tisular, Universitat Politecnica de Valencia, Spain}
\affil[3]{Computer Vision Center, Spain}
\affil[4]{Universitat Autònoma de Barcelona, Spain}
\affil[+]{these authors contributed equally to this work}

\affil[*]{pablo.hernandez-camara@uv.es}


\begin{abstract}
Contrastive language–image models such as CLIP have demonstrated remarkable generalization capabilities. 
However, how their internal visual representations evolve during training and how this evolution relates to human perception remains poorly understood. Most existing analyses characterize fully trained models, leaving the dynamics of representational biases and perceptual alignment largely unexplored. In this work, we present an epoch-by-epoch analysis of CLIP models throughout training, focusing on the evolution of texture–shape bias, alignment with human perceptual judgments, and sensitivity to image noise. Using multiple perceptual benchmarks spanning low-level image quality assessment, mid-level perceptual similarity, saliency correspondence, and noise robustness, we identify a consistent, training-stage–dependent representational transition. Early training stages exhibit strong texture bias, elevated alignment with low-level human perceptual measures, and increased sensitivity to Gaussian noise perturbations. As training progresses, this texture bias gradually diminishes in favor of more shape-based representations, coinciding with improved robustness to noise and a decline in low-level perceptual alignment. Importantly, these dynamics are consistently observed across multiple CLIP model scales, indicating that the phenomenon is not specific to a particular architecture size. Our findings provide an empirical characterization of how perceptual alignment, feature bias, and robustness co-evolve during multimodal model training. This work reveals a systematic trade-off between early low-level perceptual alignment and later robustness, offering new insights into the representational dynamics of vision–language models and their relationship to human visual processing.
\end{abstract}
\begin{document}

\flushbottom
\maketitle
%
%
\thispagestyle{empty}


\section*{Introduction}

Contrastive Language–Image Pretraining (CLIP) models, which align visual and linguistic representations using large-scale multimodal data, have established new benchmarks in zero-shot learning and transferability in computer vision \cite{radford2021learning, zhai2023sigmoid, cherti2023reproducible, chuang2025meta}. This success has motivated extensive research into the representations learned by these models \cite{kazemi2024we}, typically focusing on the properties of the final, fully trained checkpoint. Prior work has examined aspects such as classification accuracy \cite{cherti2023reproducible, qian2024online}, robustness to common image corruptions \cite{hendrycks2018benchmarking, tu2023closer, croce2025adversarially}, and alignment with human perceptual judgments \cite{geirhos2021partial}. While standard metrics such as ImageNet accuracy are usually monitored during training to verify learning progression \cite{cherti2023reproducible}, the evolution of internal feature representations throughout training remains largely underexplored. In particular, the development of fundamental representational biases and their relationship to human perception over the course of training remains an open question.

Previous studies have shown that deep neural networks exhibit systematic preferences in the visual features they exploit. Convolutional Neural Networks (CNNs) are known to exhibit a strong bias toward texture-based cues \cite{geirhos2018imagenet}, whereas Vision Transformers (ViTs) demonstrate a comparatively greater reliance on shape, though still with measurable feature biases \cite{geirhos2021partial}. Multimodal contrastive models such as CLIP, which commonly employ ViT backbones, have been shown to attenuate texture bias relative to CNNs \cite{geirhos2021partial}. Nevertheless, their representations remain substantially more texture-biased than human visual perception, which relies predominantly on shape for object recognition \cite{landau1988importance}. Importantly, existing analyses almost exclusively characterize trained model checkpoints, leaving unexamined how these biases emerge and transform during training. This leaves an important gap, as the evolution of learned visual features is likely to influence both robustness and the degree to which model representations align with the hierarchical organization of human vision. In this work, we focus on CLIP models due to the availability of dense training checkpoints, enabling an epoch-by-epoch analysis of representational dynamics.

By systematically analyzing CLIP representations across training epochs, we observe a consistent and pronounced transition in feature preferences. Early in training, the model exhibits a strong texture bias, which gradually diminishes as training progresses and representations become increasingly shape-oriented. This shift is not merely an internal representational change; it is accompanied by systematic variations in perceptual alignment and robustness-related properties. To explore the functional implications of this evolving texture–shape bias, we examine its relationship with multiple perceptual and robustness metrics. We hypothesize that early reliance on low-level, texture-based features is associated with increased alignment to early stages of human visual processing, but also with heightened sensitivity to noise. Our empirical results are consistent with this interpretation:

\begin{itemize}
    \item Early training stages exhibit pronounced texture bias alongside a peak in alignment with low-level human perceptual measures, including image quality assessment and saliency correspondence.
    \item During the same stages, the model shows increased vulnerability to Gaussian noise perturbations, reflected in larger normalized drops in classification accuracy.
    \item As training progresses and texture bias decreases in favour of more shape-based representations, noise robustness improves, while alignment with low-level human perceptual metrics declines.
\end{itemize}

Taken together, these findings suggest the presence of a shared representational transition linking texture–shape bias, perceptual alignment, and noise sensitivity during CLIP training. Crucially, we observe the same qualitative dynamics across multiple CLIP model scales (Base, Large, and Huge), indicating that this phenomenon is not confined to a specific model size. This study thus provides an empirical characterization of perceptual and representational dynamics in multimodal models, highlighting a trade-off between early low-level perceptual alignment and later robustness as training progresses. The remainder of the paper details the methodology, presents the evolution of texture–shape bias as a central organizing factor, and analyzes its relationship with human alignment and noise sensitivity.

\section*{Related work}

Understanding how representations in deep neural networks evolve during training has recently begun to attract increased attention, particularly within the broader study of representation learning dynamics. Several recent works investigate changes in learned representations throughout the training process in unimodal settings. For example, Sharon \& Dar analyzed how representations change over the course of training in deep neural networks, exploring layer-wise similarity and dynamics across architectures and optimizers, and revealing distinct training phases in vision models such as ViTs and ResNets \cite{sharon2024architecture}. Kapoor et al. study how representational alignment between networks evolves across layers and training, showing that much of the convergence occurs early and may be driven by shared input statistics and architectural biases \cite{kapoor2025bridging}. Other work investigates representational change rates during self-supervised training, observing rapid evolution in early epochs followed by stabilization \cite{Tousi2024_CCN}. While these studies provide valuable insights into how representations change, they primarily focus on unimodal models and intra-model similarity metrics, rather than on perceptual alignment with human vision or the functional consequences of these changes.

The alignment between artificial representations and human perception or neural data has also been widely studied, though predominantly for fully trained models. Prior work on human alignment examines how architectural choices, training data, or learning objectives influence correspondence with human similarity judgments, without addressing how such alignment emerges over the course of training \cite{muttenthaler2023human}. In parallel, computational neuroscience research has explored model–brain alignment using encoding models that relate neural network activations to human neural responses measured with EEG, fMRI, or MEG \cite{ferrante2024towards}. More recently, Raugel et al. investigate the evolution of alignment between self-supervised vision models and brain representations during training, showing that alignment with early sensory regions tends to emerge before higher-level correspondence \cite{demircan2024evaluating}. These findings underscore the importance of when during training alignment arises, but they focus on unimodal self-supervised models and neurophysiological measurements, rather than perceptual benchmarks and multimodal objectives.

Research on robustness, feature biases, and perceptual alignment in vision models has largely characterized static, final representations. Studies on texture–shape bias and human-like recognition highlight systematic differences between CNNs and transformer-based models, and show that multimodal contrastive objectives can mitigate texture bias relative to purely visual training \cite{muttenthaler2025aligning}. However, these analyses typically assess trained checkpoints and do not examine how feature biases, perceptual alignment, and robustness properties co-develop throughout training.

In contrast to prior work, our study provides a systematic, epoch-by-epoch characterization of how texture–shape bias, human perceptual alignment, and noise robustness jointly evolve during training in multimodal contrastive models such as CLIP. Rather than focusing solely on abstract representational similarity or neural predictivity, we directly relate representational changes to perceptual benchmarks and robustness outcomes throughout training. By demonstrating that these dynamics are consistent across multiple CLIP model scales, our work reveals a previously undocumented training-dependent trade-off between early low-level perceptual alignment and later shape-based robustness in vision–language models.

\section*{Methods}

\subsection*{Models and Training Trajectory}

We analyze the training dynamics of vision–language models based on the CLIP framework. Our primary analyses are conducted on the OpenCLIP models, for which dense training checkpoints are publicly available \cite{cherti2023reproducible}. We evaluate, without additional training or fine-tuning, all model checkpoints from epoch 0 (random initialization) through the last available checkpoint, corresponding to the full training trajectory. To assess the generality of the observed trends, we replicate all key analyses on different OpenCLIP variants (ViT-base, ViT-Large and ViT-Huge).

\subsection*{Evaluation Metrics}

To characterize how model representations evolve throughout training, we evaluate a set of complementary metrics at each training epoch. These metrics are designed to probe baseline task performance, feature bias, perceptual alignment with human vision, robustness to noise, and attentional correspondence with human saliency. All metrics are computed independently for each training epoch, enabling a direct comparison of their temporal evolution. Importantly, all evaluations are performed on pretrained checkpoints without any task-specific fine-tuning or additional supervision. This epoch-wise analysis allows us to identify consistent trends and co-evolving patterns in feature bias, perceptual alignment, robustness, and attentional focus throughout the training process. For epoch 0 (random weight initialization), results are averaged over 20 independent initializations.

\subsubsection*{Baseline Performance: Zero-Shot Classification Accuracy}

We evaluate zero-shot classification accuracy as a reference measure of task performance and learning progression. For each checkpoint, we compute zero-shot accuracy on ImageNet-1K \cite{deng2009imagenet} using the standard CLIP protocol \cite{radford2021learning}. Image embeddings are compared to text embeddings generated from class prompts of the form “An image of a \{class\}”, and predictions are obtained via maximum cosine similarity.

Zero-shot ImageNet accuracy is routinely tracked during training to verify learning progression. We report this metric following standard practice, using it as a reference indicator rather than as a central focus of the study.

\subsubsection*{Feature Bias Analysis: Texture–Shape Preference}

To quantify the model’s preference for texture versus shape cues, we evaluate texture–shape bias using the cue-conflict dataset introduced by Geirhos et al. \cite{geirhos2018imagenet}. This dataset contains images in which the global shape corresponds to one object category while the local texture corresponds to another.

For each conflict image, predictions are obtained following the standard Geirhos et al. evaluation procedure\cite{geirhos2018imagenet}, in which model outputs over ImageNet classes are mapped to the corresponding shape and texture categories. An image is counted as texture-biased or shape-biased depending on which cue matches the predicted category. Texture bias is then quantified as the proportion of images for which the model prediction aligns with the texture rather than the shape. This procedure is repeated independently for each training epoch, allowing us to track the evolution of texture–shape preference throughout training.

\subsubsection*{Robustness to Noise Perturbations}

Robustness to noise is assessed by measuring the model’s sensitivity to Gaussian noise corruptions. We evaluate zero-shot classification accuracy on the ImageNet-1K-C dataset \cite{hendrycks2019robustness}, considering Gaussian noise at five predefined severity levels.

For each training epoch, accuracy is computed separately for each severity level and then averaged. The normalized accuracy drop is defined as the relative decrease between clean ImageNet accuracy and the mean accuracy across noise levels. This metric captures the model’s overall sensitivity to noise perturbations and enables consistent comparisons across training epochs.

\subsubsection*{Perceptual Alignment with Human Vision}

\textbf{Low-Level Perceptual Alignment}

Low-level perceptual alignment is evaluated using the TID2013 \cite{ponomarenko2015image} and KADID-10K \cite{kadid10k} image quality assessment databases. These datasets consist of reference images and corresponding distorted versions, along with human Mean Opinion Scores (MOS) reflecting perceived image quality. The distortions primarily affect low-level image statistics, such as noise, blur, and compression artifacts.

For each epoch, we compute image embeddings for reference and distorted image pairs and measure their cosine similarity in the model’s embedding space. The resulting similarity scores are then correlated with the human MOS values using Spearman correlation. This correlation serves as a measure of how well model representations align with human judgments of low-level perceptual quality.

\textbf{Mid-Level Perceptual Alignment}

To assess perceptual alignment beyond low-level distortions, we evaluate the model using the NIGHTS perceptual database \cite{fu2023dreamsim}, which focuses on mid-level perceptual differences such as object pose, color, and quantity. Each sample consists of a reference image and two distorted versions, along with human judgments indicating which distorted image is perceptually closer to the reference.

For each training epoch, we compute embedding similarities between the reference image and each distorted image. The model is considered correct if it assigns higher similarity to the distorted image preferred by humans. Accuracy is computed as the proportion of trials in which the model’s preference matches human judgments. This metric provides a measure of mid-level perceptual alignment across training.

\subsubsection*{Attentional Alignment with Human Saliency}

To examine the correspondence between model attention and human visual attention, we evaluate saliency alignment using human saliency maps from the MIT1003 dataset \cite{Judd_2009}. Model attention maps are obtained using the attention roll-out method for vision transformers, which aggregates attention weights across layers to produce a spatial attention distribution over the input image.

For each image, we compute the correlation between the model’s attention roll-out map and the corresponding human saliency map. This correlation serves as a measure of attentional alignment and allows us to track how correspondence with human visual attention evolves over the course of training.

\section*{Results}

\subsection*{Baseline Learning and Feature Bias Evolution}

\begin{figure}[ht]
\centering
\includegraphics[width=\linewidth]{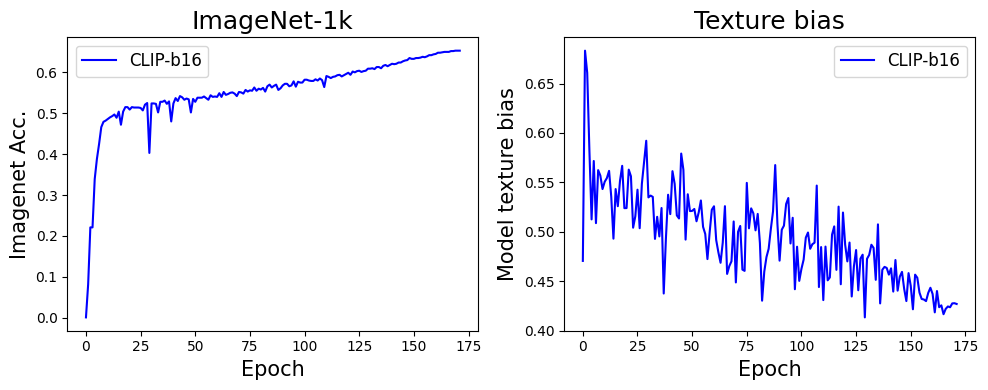}
\caption{\textbf{Left:} ImageNet-1k accuracy for CLIP-base16 during the training. \textbf{Right:} CLIP-base16 texture bias based on cue-conflict images.}
\label{fig_imagenet_texture_clip_base}
\end{figure}

To verify the model's learning progress, we report zero-shot classification accuracy on ImageNet-1K as shown in figure \ref{fig_imagenet_texture_clip_base} left plot. As expected, classification accuracy increases steadily throughout training, confirming that the model is learning meaningful representations across training epochs. These accuracy measures serve as a reference for task performance and provide insight into the model’s general learning trajectory.

We evaluate the model’s preference for texture versus shape using the cue-conflict texture-shape bias images \cite{geirhos2018imagenet}. Figure \ref{fig_imagenet_texture_clip_base} right plot shows how the model initially exhibits a strong texture bias, which progressively decreases over the course of training. This shift reflects the model’s transition from reliance on low-level texture features to more abstract shape-based representations, marking a key aspect of the model’s feature evolution.

\subsection*{Alignment and Robustness Dynamics}

\begin{figure}[ht]
\centering
\includegraphics[width=\linewidth]{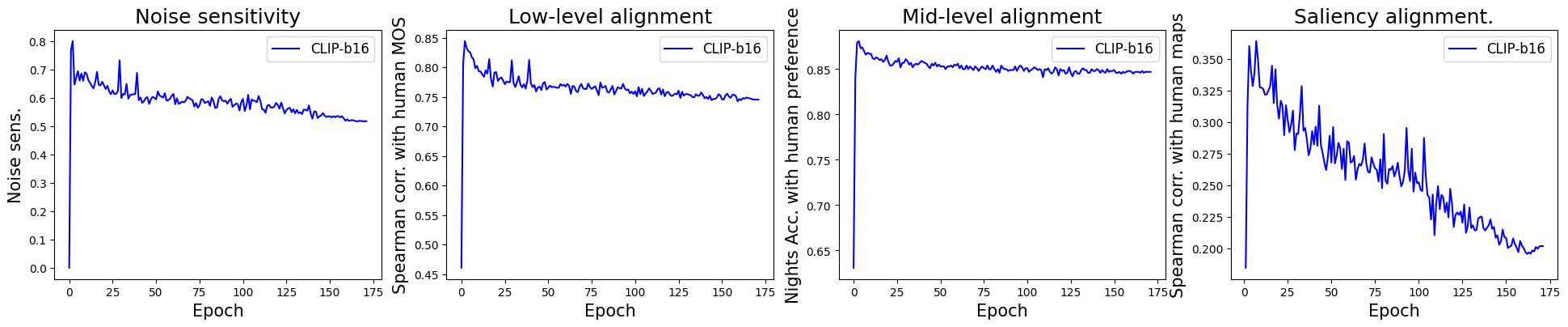}
\caption{Evolution of CLIP-base16 on four different datasets during its training. From left to right, it shows: 1. Noise sensitivity: Measured as the normalized accuracy drop on ImageNet-1k-C gaussian noise corrupted images. 2. Low-level alignment: Mean model correlation with human mean opinion score on both TID2013 and KADID-10K. 3. Mid-level alignment: Model accuracy with human preferences on the NIGHTS perceptual database. 4. Saliency alignment: Model correlation with human saliency maps from .}
\label{fig_consequences_clip_base}
\end{figure}

Figure \ref{fig_consequences_clip_base} presents the model robustness to noise perturbation and the alignment with human low-level, mid-level and saliency perceptual judgments. 

Regarding noise sensitivity, the model exhibits the greatest sensitivity to noise early in training, when the texture bias is most pronounced. As training advances and the model shifts toward shape-based features, noise sensitivity decreases, demonstrating a trade-off between early perceptual alignment and later robustness. 

When focusing on the low- and mid-level alignment, we observe that the alignment peaks in the early epochs before gradually declining, which corresponds to the initial high texture bias observed during this period. This decrease in alignment coincides with the model transition from texture-based features to more shape-oriented representations as training progresses.

Finally, regarding saliency alignment, we also observe that the attentional alignment peaks in the early epochs, again when the model is more texture-biased.

\subsection*{Quantitative Correlation and Generalizability
Correlation Analysis}

\begin{table}[ht]
\centering
\begin{tabular}{|c|c|c|c|c|c|c|}
\hline
 & ImageNet-1k & Texture-bias & Noise sens. & Low-level align. & Mid-level align. & Saliency align. \\
\hline
ImageNet-1k & 1 & -0.794 & -0.904 & -0.890 & -0.806 & -0.919 \\
\hline
Texture-bias &  & 1 & 0.780 & 0.753 & 0.695 & 0.765 \\
\hline
Noise sens. &  &  & 1 & 0.891 & 0.788 & 0.876 \\
\hline
Low-level align. &  &  &  & 1 & 0.786 & 0.865 \\
\hline
Mid-level align. &  &  &  &  & 1 & 0.863 \\
\hline
Saliency align. &  &  &  &  &  & 1 \\
\hline
\end{tabular}
\caption{\label{tab_correlations_clip_base}Spearman correlations between the different analyzed metrics during the training for the CLIP-base16 model.}
\end{table}

Table \ref{tab_correlations_clip_base} presents the Spearman correlation matrix, quantifying the relationships between ImageNet1k accuracy, texture bias, noise sensitivity, low- and mid-level human alignment and saliency alignment throughout training. Strong correlations between these metrics suggest that these metrics share a common developmental trajectory, with early texture-dominated representations associated with higher low-level alignment and increased noise sensitivity, which later give way to shape-based representations and improved robustness.

\begin{figure}[ht]
\centering
\includegraphics[width=\linewidth]{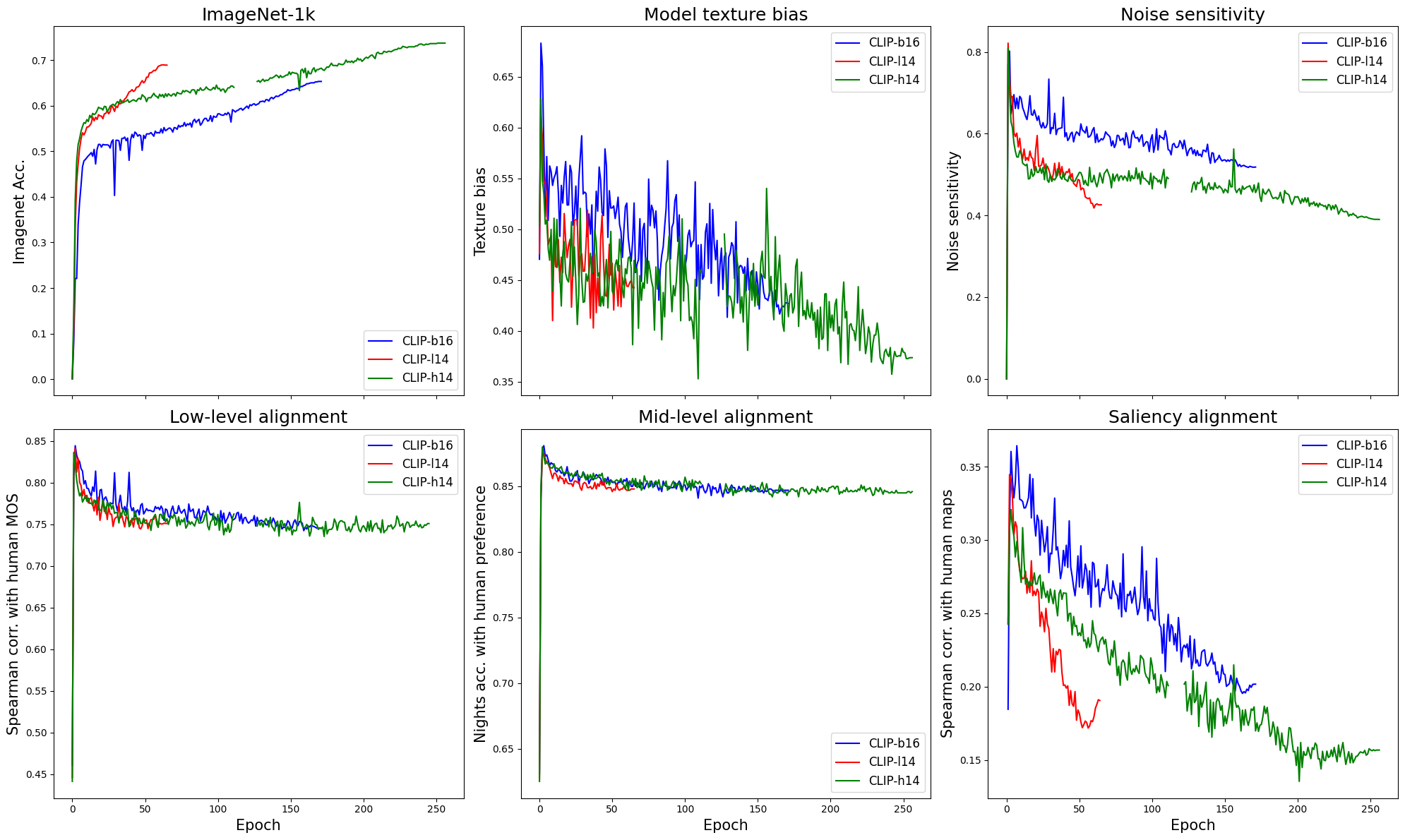}
\caption{Evolution of the analyzed metrics for three different CLIP scales (base, large, huge).}
\label{fig_all_model_scales}
\end{figure}

Finally, to assess the generalizability of these dynamics, we present a summary of results across three different model sizes (Base, Large, and Huge). Figure \ref{fig_all_model_scales} shows that the same patterns of texture bias reduction, perceptual alignment shift, and robustness improvement hold consistently across all three model scales. This indicates that the observed dynamics are not specific to a single model size, but rather reflect a general characteristic of the CLIP training process.

\section*{Discussion}

In this work, we provide an empirical analysis of how visual representations in CLIP models evolve throughout training, revealing a consistent transition from low-level, texture-dominated representations toward more abstract, shape-based features. This transition is observed across multiple complementary metrics, including texture–shape bias, perceptual alignment with human judgments, noise sensitivity, and attentional correspondence with human saliency. Importantly, these dynamics are consistent across model scales, indicating that they reflect a general property of CLIP training rather than a size-specific effect.

By examining representations throughout training rather than only at convergence, our results extend prior studies of feature bias and human alignment and show that these properties are strongly training-stage dependent. Early in training, CLIP exhibits a pronounced texture bias, high alignment with low-level human perceptual judgments, and increased sensitivity to noise. As training progresses, representations gradually shift toward shape-based abstractions that are more robust to noise but less aligned with low-level perceptual measures. This co-evolution indicates that perceptual alignment and robustness are not static characteristics of the model, but change systematically with the underlying feature representations learned during training.

A central insight emerging from this analysis is a trade-off between early perceptual alignment and later robustness. Metrics associated with early stages of human vision, such as image quality assessment and saliency correspondence, peak when the model relies most strongly on texture-based cues, but this reliance is also associated with greater vulnerability to noise perturbations. As texture bias decreases and shape-based representations become more prominent, robustness improves while alignment with low-level perceptual metrics declines. Rather than reflecting a limitation of the model, this pattern suggests that the trade-off is a natural consequence of representational changes induced by contrastive multimodal training.

These findings have practical implications for the use and development of vision–language models. Tasks that depend on sensitivity to low-level visual attributes may benefit from representations obtained at earlier training stages, whereas applications prioritizing robustness and semantic abstraction may prefer later-stage representations. More broadly, our results suggest that training objectives implicitly regulate the balance between perceptual alignment and robustness, motivating future work on training strategies that explicitly modulate this transition. Extending this analysis to other multimodal architectures, training objectives, and supervision paradigms may further clarify how perceptual and robust representations emerge during learning.

\section*{Conclusions}

We have shown that CLIP training is characterized by a systematic transition from low-level, texture-based representations toward more abstract, shape-based features. This transition is accompanied by coordinated changes in human perceptual alignment, noise sensitivity, and attentional correspondence, revealing a systematic trade-off between early perceptual sensitivity and later robustness. By demonstrating that these dynamics are consistent across model scales, our work highlights representational evolution as a fundamental aspect of multimodal model training.

These findings emphasize the importance of considering training dynamics—not only final model checkpoints—when evaluating alignment with human perception and robustness. Understanding how and when specific representational properties emerge provides new opportunities to tailor models to perceptual or robustness-driven applications, and offers a framework for future research on controlling representational trade-offs in vision–language models.

\bibliography{sample}








\end{document}